\documentclass[runningheads]{llncs}
\usepackage[pdftex]{graphicx}
\usepackage[pdftex]{color}
\usepackage{comment}
\usepackage{amsmath,amssymb} 
\usepackage{color}
\usepackage{multirow}
\usepackage{subfigure}
\usepackage{bm}
\usepackage{ulem}

\makeatletter
\newcommand{\figcaption}[1]{\def\@captype{figure}\caption{#1}}
\newcommand{\tblcaption}[1]{\def\@captype{table}\caption{#1}}
\makeatother

\DeclareMathOperator*{\argmax}{arg\,max}

\begin{document}
\pagestyle{headings}
\mainmatter
\def\ECCVSubNumber{1489}  

\title{Weakly-Supervised Cell Tracking\\
via Backward-and-Forward Propagation} 

\titlerunning{Weakly-Supervised Cell Tracking via Backward-and-Forward Propagation}
%
\author{Kazuya Nishimura\inst{1}\and
Junya Hayashida\inst{1}\and
Chenyang Wang\inst{2}\and
Dai Fei Elmer Ker\inst{2}\and
Ryoma Bise\inst{1}
}
\authorrunning{K. Nishimura, J. Hayashida, et al.}
%
\institute{Kyushu University, Fukuoka, Japan \\
\email{\{kazuya.nishimura,bise\}@human.ait.kyushu-u.ac.jp}
\and
The Chinese University of Hong Kong, Hong Kong}
\maketitle

\begin{abstract}
We propose a weakly-supervised cell tracking method that can train a convolutional neural network (CNN) by using only the annotation of ``cell detection'' ({\it i.e.}, the coordinates of cell positions) without association information, in which cell positions can be easily obtained by nuclear staining.
First, we train co-detection CNN that detects cells in successive frames by using weak-labels.
Our key assumption is that co-detection CNN implicitly learns association in addition to detection.
To obtain the association, we propose a backward-and-forward propagation method that analyzes the correspondence of cell positions in the detection maps output of co-detection CNN.
Experiments demonstrated that the proposed method can associate cells by analyzing co-detection CNN. Even though the method uses only weak supervision, the performance of our method was almost the same as the state-of-the-art supervised method.
Code is publicly available in \url{https://github.com/naivete5656/WSCTBFP} .
\keywords{Cell tracking, weakly-supervised learning, multi-object tracking, cell detection, tracking, weakly-supervised tracking}
\end{abstract}

\section{Introduction}
\vspace{-2mm}
Cell behavior analysis plays an important role in biology and medicine.
To create quantitative cell-behavior metrics, cells are often captured with time-lapse images by using phase-contrast microscopy, which is a non-invasive imaging technique, and then hundreds of cells over thousands of frames are tracked in populations. However, it is time-consuming to track a large number of cells manually. Thus, automatic cell tracking is required.

Cell tracking in phase-contrast microscopy has several difficulties compared with general object tracking.
First, cells have similar appearences and their shapes may be severely deformed.
Second, cells often touch each other and have blurry intercellular boundaries. 
Third, a cell may divide into two cells (cell mitosis); this is very different from general object tracking. 
These aspects make it difficult to track cells by using only shape similarity and proximity of cells.

To address such difficulties, the positional relationship of nearby cells is important information to identify the association.
The recently proposed CNN-based methods that use such context~\cite{payer2018instance,hayashida2019cell} have outperformed the conventional image-processing-based methods. However, learning-based methods require enough training data including individual cell positions in each frame and their correspondences in successive frames ({\it i.e.,} cell location and motion).
In addition, the annotation process may not be a one-time event due to the variety of cell types and culturing environments ({\it e.g.,} growth-factors, type of microscope). 
Since the apparent shape of a cell and its behaviors may often change depending on such conditions, we usually have to prepare a training dataset for each individual case.

\begin{figure}[t]
    \centering
    \includegraphics[width=0.9\linewidth]{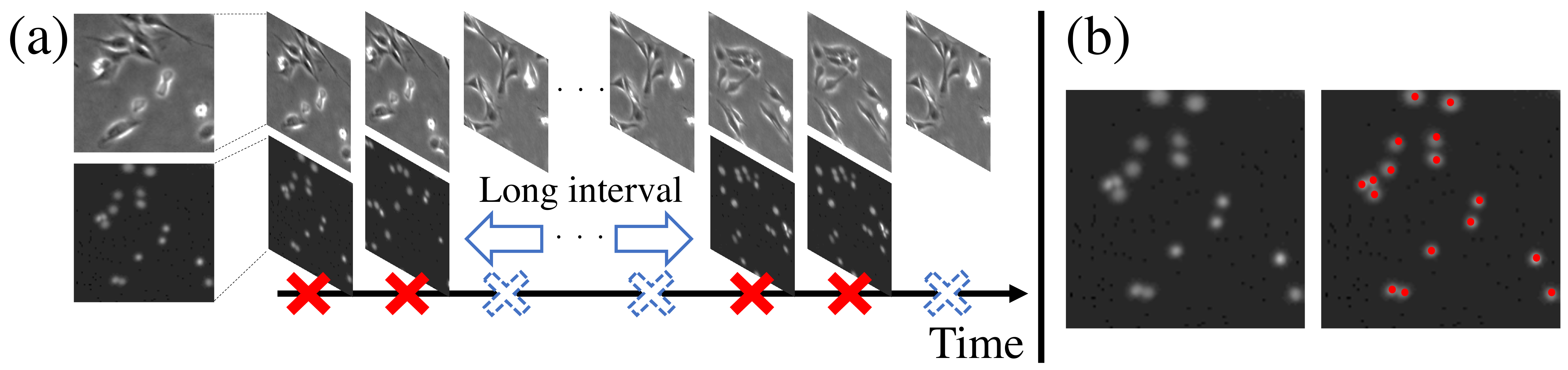}
    \vspace{-3mm}
    \caption{(a) \textbf{Top}: Phase-contrast image sequence. \textbf{Bottom}: Fluorescent images with the nucleus stain cell; these were only captured with longer intervals (red crosses) due to phototoxicity. Fluorescent images were not captured on the white crosses. 
    (b) Rough centroid positions (in red) can be easily identified in fluorescent images. 
    }
    \label{fig:phase_and_fluorescent}
    \vspace{-5mm}
\end{figure}

On the other hand, there are invasive imaging techniques such as fluorescent imaging to facilitate observation of cells.
If we can obtain fluorescent images showing cells whose nuclei are stained (Fig. \ref{fig:phase_and_fluorescent}) in addition to the phase-contrast images, the rough centroid positions can be easily detected by using simple image processing techniques. 
However, because fluorescent imaging damages cells, these images can be only captured for training, not for testing.
Moreover, fluorescent images cannot be captured frequently over a long period, since phototoxicity may affect the shapes and migration of the cells.
Instead, we can capture fluorescent images only several times in enough long period (Fig. \ref{fig:phase_and_fluorescent}) since cells can recover from the damage during the non-invasive imaging period.
From such sequences, we can automatically obtain point labels for detection~\cite{nishimura2019weakly}. Although these labels do not include the correspondence information between frames, they can be considered as weak-labels for the tracking task. 

\begin{figure*}[t]
    \centering
    \includegraphics[width=0.88\linewidth]{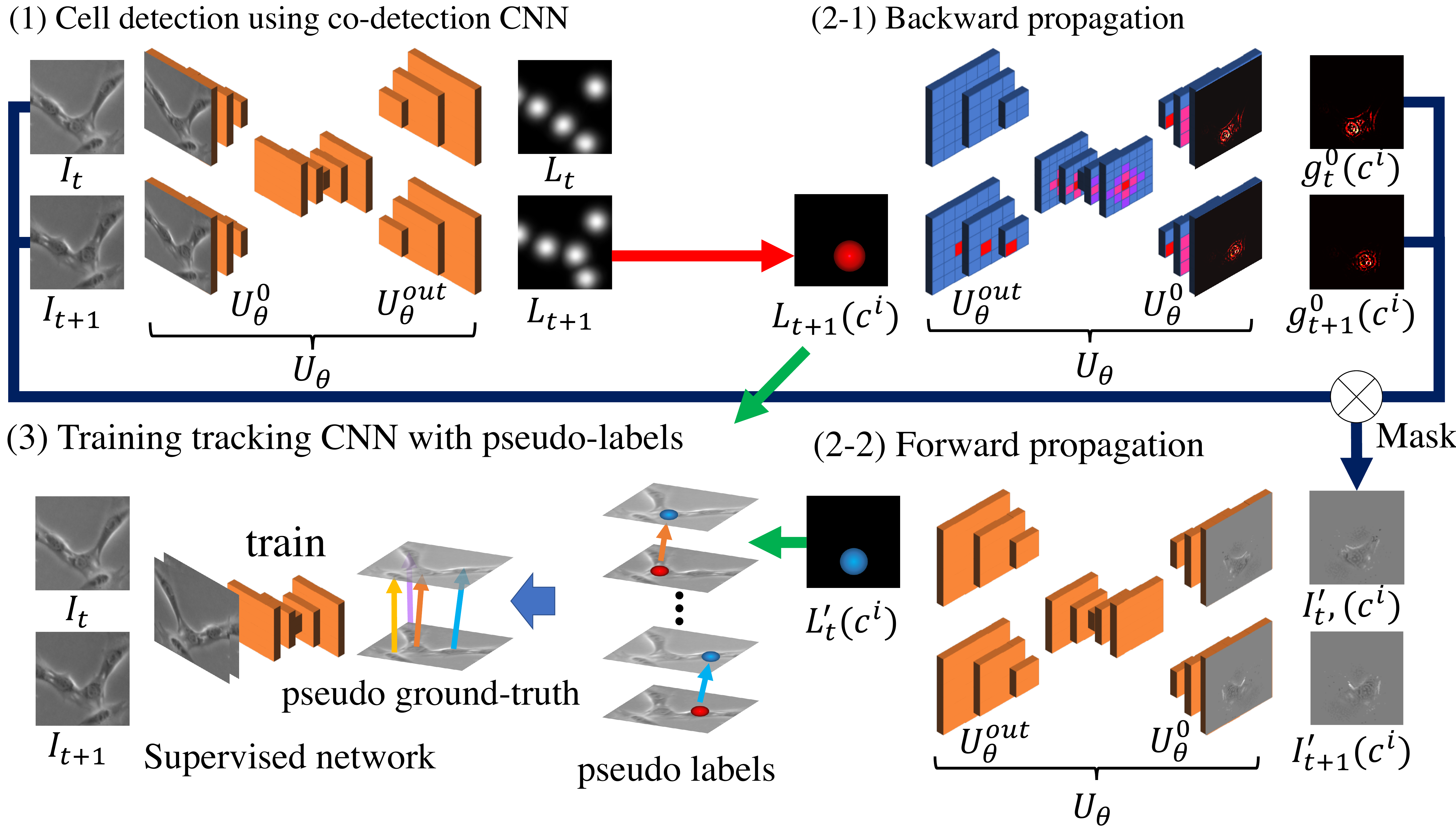}
    \vspace{-6mm}
    \caption{Overview of our method. (1) co-detection CNN $U_{\theta}$ estimates the position likelihood maps for two successive frames.
    (2-1) Backward-propagation estimates relevance maps $g^0_{t}(c^i)$, $g^0_{t+1}(c^i)$ of the cell region of interest $c^i$ (red).
    (2-2) Forward-propagation estimates the cell position likelihood map $L'_t(c^i)$ (blue) with inputting the masked images $I'_{t}(c^i)$, $I'_{t+1}(c^i)$ which are generated using $g^0_{t}(c^i)$, $g^0_{t+1}(c^i)$. The pseudo-labels are generated using this estimated regions. 
    (3) The tracking CNN is trained using the pseudo-labels.}
    \label{fig:method_overview}
    \vspace{-2mm}
\end{figure*}

In this paper, we propose a weakly supervised cell tracking that can obtain the correspondences from training data that are supposed to be used in a detection task but not for tracking.
In order to obtain the association information, we designed a method that has three steps as shown in Fig. \ref{fig:method_overview}:
(1) Our co-detection CNN is trained to detect cells in successive frames by using the rough cell centroid positions, which are weak-labels for the tracking task but it can be used as supervision for detection.
Our key assumption is that co-detection CNN implicitly learns the association.
(2) The proposed method performs backward-and-forward propagation to extract associations from co-detection CNN without any ground-truth. When we focus on a particular detection response in the output layer $L_{t+1}$ ({\it e.g.,} the red region in the left image of Fig. \ref{fig:method_overview} (2-1)), the association problem can be considered to be one of finding the position corresponding with the cell of interest (blue region in Fig. \ref{fig:method_overview} (2-2)) from $L_{t}$. 
The backward-and-forward propagation can obtain association.
(3) Using the detection results (1) and association results (2), we can generate the pseudo-training data for the tracking task.
We train the cell tracking method \cite{hayashida2019cell} with pseudo-training data and a masked loss function that ignores the loss from the false-negative regions, in which we can know the false-negative regions where the cell of interest cannot be associated with any cells in the second step.
It is expected that the trained tracking network has better tracking performance compared with the pseudo-labels.

Our main contributions are summarized as follows:
\begin{itemize}
\vspace{-1mm}
    \item We propose a weakly-supervised tracking method that can track multiple cells by only using training data for detection. 
    Our method obtains cell association information from co-detection CNN. The association information is used as pseudo-training data for cell tracking CNN. 
    \item We propose a novel network analysis method for determining corresponding positions in two maps output from multi-branch network. Our method can extract the positional correspondences of cells from two successive frames by analyzing co-detection CNN. 
    \item We demonstrated the effectiveness of our method using open data and realistic data. 
    In realistic data, we do not use any human annotations.
    Our method outperformed current methods that do not require training data. In addition, even though the method uses only weak supervision, the performance of our method was almost the same with the state-of-the-art supervised method.
\end{itemize}

\section{Related work}
\noindent {\bf Cell tracking:}
Many cell tracking methods have been proposed, which is particle filters~\cite{okuma2004boosted,smal2006bayesian}, active contour ~\cite{li2008cell,wang2007cell,yang2005cell,zhou2019joint}, and detection-and-association~\cite{KanadeT11,BiseR2009,bise2011reliable,bise2013,schiegg2013conservation,su2013cell,zhou2019joint}.
The detection-and-association methods, which first detect cells in each frame and then solve associations between successive frames, are the most popular tracking paradigm due to the good quality of detection algorithms that use CNNs in the detection step~\cite{rempfler2017cell,rempfler2018tracing,akram2016joint,lux2019dic}.
To associate the detected cells, many methods use hand-crafted association scores based on proximity and shape similarity~\cite{KanadeT11,zhou2019joint,bise2011reliable,schiegg2013conservation,su2013cell}.
To extract the similarity features from images, 
Payer {\it et al.}~\cite{payer2018instance} proposed a recurrent hourglass network that not only extracts local features but also memorizes inter-frame information.
Hayashida {\it et al.}~\cite{hayashida2019cell,hayashida2019mpm} proposed a cell motion field that represents the cell association between successive frames and it can be estimated by a CNN. These methods outperform ones that use hand-crafted association scores. However, they require sufficient training data for both detection and association.

\noindent {\bf Unsupervised or weakly-supervised tracking for general objects:} Recently, several unsupervised or weakly-supervised tracking methods have been proposed. 
To track a single object, correspondence learning have been proposed with several weakly-supervision~\cite{zhong2014visual} or unsupervised scenarios~\cite{vondrick2018tracking,wang2019learning,wang2019unsupervised}. Zhong {\it et al.}~\cite{zhong2014visual} proposed a tracking method that combines the outputs of multiple trackers to improve tracking accuracy in order to address noisy labels (weak labels). The weak label scenario is different from ours.
These methods assumed for tracking a single object, and thus these are short to our problem.
Several methods have been proposed for multi-object tracking~\cite{nwoye2019weakly,huang2020multiple}.
Nwoye {\it et al.}~\cite{nwoye2019weakly} proposed a weakly-supervised tracking of surgical tools appearing in endoscopic video.
The weak label in this case is a class label of the tool type, and they assumed that one tool appears for each tool type even though several different types of tools appear at a frame.
Huang {\it et al.}~\cite{huang2020multiple} tackled a similar problem of semantic object tracking.
He {\it et al.}~\cite{he2019tracking} proposed an unsupervised tracking-by-animation framework.
This method uses shape and appearance information for updating the track states of multiple-objects. It assumes that the target objects have different appearances. 
The above methods may become confused if there are many similar appearance objects in the image; such is the case in cell tracking.

\noindent {\bf Relevant pixel analysis:}
Visualization methods have been proposed for analyzing relevant pixels for classification in CNNs~\cite{smilkov2017smoothgrad,bach2015pixel,montavon2017explaining,springenberg2015striving,zhou2016learning,selvaraju2017grad,chattopadhay2018grad,zhang2018top,kindermans2017learning}.
Layer-wise relevance propagation (LRP)~\cite{bach2015pixel,montavon2017explaining} and guided backpropagation~\cite{springenberg2015striving} back-propagate signals from the output layer to the input layer on the basis of the weights and signals in the forward-propagation for inference.
Methods based on class activation mapping (CAM)~\cite{zhou2016learning,selvaraju2017grad,chattopadhay2018grad}, such as Grad-CAM~\cite{chattopadhay2018grad}, 
produces the relevance map from CNN using the semantic features  right before the fully connected layer for classification.
There are several methods that uses such backward operation in a network for instance segmentation~\cite{nishimura2019weakly} and object tracking~\cite{li2019gradnet}.
For example, Li {\it et al.}~\cite{li2019gradnet} propose a Gradient-Guided Network (GradNet) for a single object tracking that exploits the information in the gradient for template update. Although this method uses the backward operation for guided calculation, the purpose is totally different from ours (analysis of the relevance of the two output layers).
These methods assume that they analyze the relationship between the input and output layers but not for two output layers in a multi-branch network.

Unlike the above methods, our method can obtain correspondences between objects in successive frames without training data for the association by analyzing the co-detection CNN, in spite of the challenging conditions wherein many cells having similar appearances migrate.

\vspace{-2mm}
\section{Weakly-supervised cell tracking}
\vspace{-2mm}
\subsection{Overview}
Fig.~\ref{fig:method_overview} shows an overview of the proposed method.
The method consists of three parts: 1) cell detection using co-detection CNN that jointly detects cells at successive frames using weak labels (cell position label): it is expected to implicitly learn not only cell localization but also the association between the frames; 2) backward-and-forward propagation for extracting the association information from co-detection CNN: we generate the pseudo labels for training a tracking network so that its precision is high enough although it may contain some false-negatives; and 3) Training the cell tracking network using pseudo-labels made by the step two: we introduce the masked loss to ignore such false-negatives.
The details of each step are explained as follows.

\begin{figure}[t]
    \centering
    \includegraphics[width=0.7\linewidth]{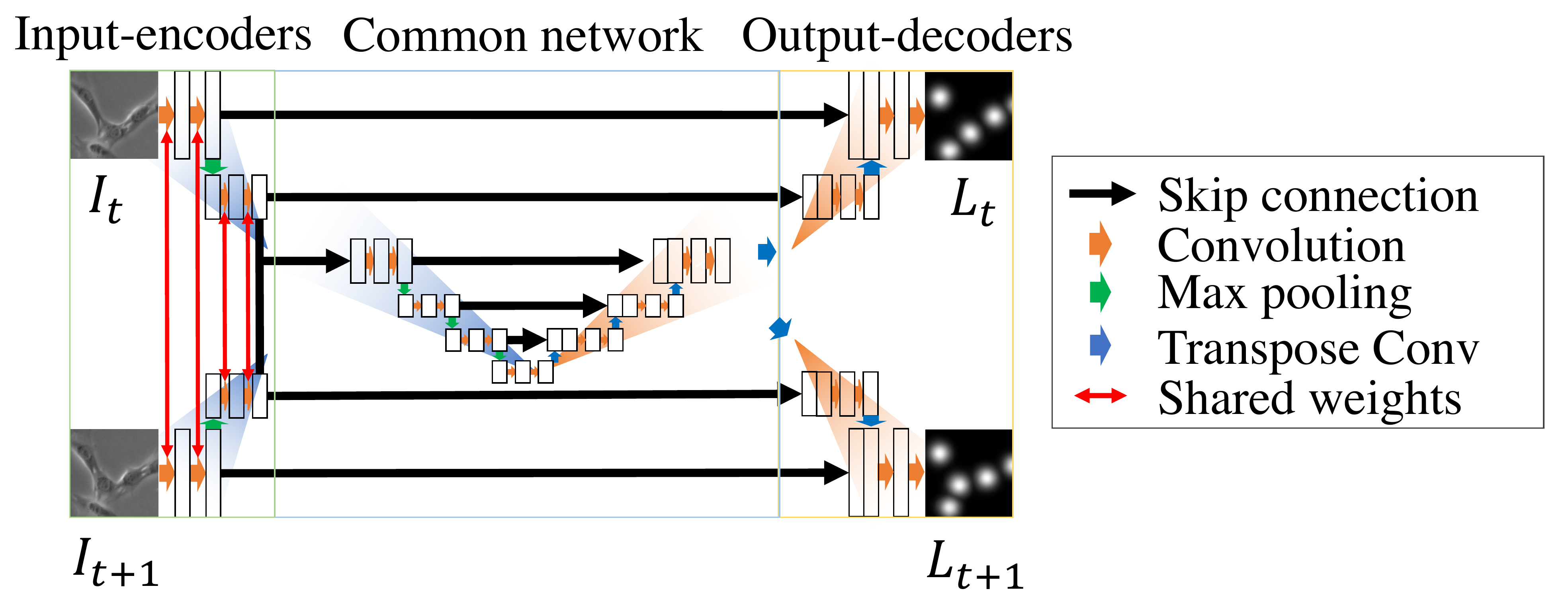}
    \vspace{-3mm}
    \caption{Architecture of co-detection CNN.}
    \label{fig:co-detection_Network}
    \vspace{-4mm}
\end{figure}

\subsection{Co-detection CNN}
\label{sec:co-detection}
In the co-detection task~\cite{bao2012object}, the detection results in frame $t$ can facilitate to detect the corresponding cell in frame $t+1$ and vice versa. Based on this key observation, we designed co-detection CNN $U_{\theta}$ for jointly detecting cells at the successive frames, in which $\theta$ indicates the network parameters.
In our problem setup, a set of the cell position coordinates in successive frames are obtained from the fluorescent images as training data, but the nuclei position may shift from the ground-truth of the centroid position. Therefore, we follow the cell detection network~\cite{nishimura2019weakly} that mitigates this gap by representing cell positions as a position likelihood map.
The ground-truth of the map can be automatically generated from the rough cell centroid position obtained from the fluorescent images, where a given cell position becomes an intensity peak and the intensity value gradually decreases away from the peak in accordance with a Gaussian distribution \cite{nishimura2019weakly}.
In contrast to \cite{nishimura2019weakly} (U-Net~\cite{ronneberger2015u} architecture), our network has two input-encoders, a common network, and two output-decoders to simultaneously estimates the detection results in successive frames as shown in Fig. \ref{fig:co-detection_Network}.
The two input-encoders have shared weights, and these extract the cell appearance features from the inputted successive images $I_t$, $I_{t+1}$.
The features are concatenated and input into a common network that has a U-Net architecture.
We consider that the common network performs co-detection and it implicitly learns the cell association. Finally, the output-decoders decode the extracted co-detection features into the cell position likelihood maps; the layers of the input and output networks have skip connections to adjust the local positions similar to U-Net.
The loss function $Loss_{CD}$ for co-detection CNN is the sum of the Mean Square Errors (MSE) of the likelihood maps of the two frames:
\begin{equation}
    Loss_{CD} = \\
    MSE(L_t - \hat{L}_t) + MSE(L_{t+1} - \hat{L}_{t+1}),
\end{equation}
where $\hat{L}_t$, $\hat{L}_{t+1}$ are the ground-truths of the cell position likelihood map of each frame and $L_t$, $L_{t+1}$ are the estimated maps. In the inference, the peaks in the estimated map are the detected cell positions.

\vspace{-2mm}
\subsection{Backward-and-Forward propagation}
\label{sec:BFProp}
\vspace{-2mm}
Next, in accordance with our assumption that co-detection CNN implicitly learns the association, we extract the cell association information from co-detection CNN $U_{\theta}$. 
Here, we will focus on a particular detection response in the output layer $L_{t+1}$ ({\it e.g.,} the red regions in Fig. \ref{fig:method_overview} (2-1)). The association problem can be considered to be one of finding the position (blue region) corresponding to the cell of interest from $L_{t}$.
We propose the following backward-and-forward propagation for this task.

\vspace{-3mm}
\subsubsection{Backward propagation:}
\label{sec:BProp}
Fig.~\ref{fig:method_overview} (2-1) illustrates the backward-propagation process on $U_{\theta}$ for the cell of interest $c^i$ that is selected from frame $t + 1$. 
In this step, we extract the relevance maps that are expected to relevant to producing the detection response of interest by using guided backpropagation (GB)~\cite{springenberg2015striving}.
For this process, we modified the weakly-supervised instance segmentation method proposed by Nishimura~\cite{nishimura2019weakly}, which extracts individual relevant cell regions of a particular cell in U-Net for a single image.
Different from \cite{nishimura2019weakly}, in our case, two relevance maps $g^0_t(c^i)$, $g^0_{t+1}(c^i)$ are extracted from a single output $L_{t+1}(c^i)$.

The GB back propagates the signals from the output layer $U^{out}_{\theta}$ to the input layer $U^{0}_{\theta}$ by using the trained parameters (weights) $\theta$ in the network. 
In our method, to obtain the individual relevant cell regions of a particular cell, we first initialize the cell position likelihood map $L_{t+1}(c^i)$ for each cell of interest $c^i$, in which all regions outside the cell region substitute 0 (Fig.~\ref{fig:method_overview} (2-1)). 
The region within radius $r$ from the coordinate of the cell of interest $c^i$ is defined as $S(c^i)$.
Then, the relevant pixels of each cell of interest were obtained by back-propagation from $S(c^i)$.
The backpropagating signals are propagated to both layers at the branch of the input-encoders. The red nodes of the intermediate layers in the network in Fig.~\ref{fig:method_overview} (2-1) show the illustration of the back-propagation process.
This backward process is performed for each cell $i=1,...,N$, where $N$ is the number of cells.

It is expected that the corresponding cell regions have positive values in the relevance maps.
However, regions of the outside the cell of interest may also have values in the relevance map. 
This adversely affects the process of extracting the cell association. 
Therefore, we compare the pixel values of $g^0_{p,t}(c^i)$ ($i=\{1,...,N\}$) for all cells, where pixel $p$ corresponds to $c^i$ if it takes the maximum value among the cells 
The maximum projection of the $p$-th pixel for $c^i$ at frame $t$ can be formalized as:

\begin{eqnarray}
    g'_{p, t}(c^i) &=& \psi_p(t, i, \argmax_{k} g_{p, t}^0 ( c^k )),\\
    \psi_p(t, i, k) &=&
    \begin{cases}
        {g}^{0}_{p, t}(c^i)&if ~(k=i), \\
        0&\mbox{otherwise},
    \end{cases}
\end{eqnarray}\label{eq:indivisual region propagation}
where, $p$ is the $p$-th pixel on the relevance map.
$g'_{t+1}(c^i)$ is calculated by same manner.
By applying maximum projection to all cells, we get the maximum projection relevance maps $g'_{t}(c^i)$ and $g'_{t+1}(c^i)$ for each cell.

\vspace{-3mm}
\subsubsection{Forward propagation:}
\label{sec:FProp}
This step estimates the corresponding cell position likelihood map $L'_{t}(c^i)$ by using the relevance maps  $g'_{t}(c^i)$ and $g'_{t+1}(c^i)$ (Fig.~\ref{fig:method_overview} (2-2)).
The high value pixels in $g'_{t}(c^i)$, $g'_{t+1}(c^i)$ show the pixels that contribute to detect $c^i$.
It indicates that co-detection CNN $U_{\theta}$ is able to detect the corresponding cell position from only these relevant pixels in the input images.

We generate masked images  $I'_{t}(c^i)$, $I'_{t+1}(c^i)$ that only have values at the high relevance pixels of only the cell of interests $c^i$.
In order to generate it, we first initialize the images so that it has the background intensities of the input image.
Then, we set the pixel values in the initialized image as the input image intensity if the value of the $p$-th pixel of  $g'_{t}(c^i)$ is larger than a threshold $th$.
\begin{equation}
    I'_{p,t}(c^i) = 
    \begin{cases}
        I_{p,t}(c^i) &\mbox{if} ~g'_{p,t}(c^i) > th,\\
        B_{p, t}(c^i)   &\mbox{otherwise},
    \end{cases}
    \label{eq:forwardmask}
\end{equation}
where the background image at frame t $B_t$ is estimated by using quadratic curve fitting~\cite{yin2012understanding} and $p$ is the p-th pixel on $I'$.
$I'_{t+1}(c^i)$ is also made by the same manner.
The corresponding cell position likelihood map $L'_{t}(c^i)=U_{\theta}(I'_t(c^i),I'_{t+1}(c^i))$ can be obtained by inputting the masked images $I'_{t}(c^i)$, $I'_{t+1}(c^i)$ to $U_{\theta}$.

The estimated map $L'_{t}(c^i)$ indicates the $i$-th detection response at $t$ that corresponds to the detected cell at frame $t$.
It is expected that the high intensity region in $L'_{t}(c^i)$ is expected to appears on the same region of either cell detection result in $L_{t}$.
To obtain the cell position at $t$ corresponding to the $i$-th cell position at $t+1$, we perform one-by-one matching by using linear programming between these two maps, in which we use a simple MSE of intensities for the matching score.
This one-by-one matching may correspond either of the cell position even if the estimated response signal is too small.
We omit the low confidence associations in order to keep the precision of the pseudo-labels high enough. 
If the matching score is less than the threshold $th_{conf}$, we define it as the low confidence association.
By omitting low confidence associations,  the result includes some false-negative.
One of the interesting points is that we can know where the low confidence region at $t+1$ is since we explicitly give the region of the cell of interest $S_{t+1}(c^i)$. 
If the cell $c^i$ is not associated with any cell, we add the pixels of the cell $S_{t+1}(c^i)$ to the set of unassociated cell region $\mathbf{\Gamma}$.
Finally, we obtain the cell position and association in successive frames and there will use as pseudo-labels in the next step.


\subsection{Training tracking CNN using pseudo-labels}
\label{sec:train with masked}
It is known that the generalization performance and estimation speed can be improved by training a CNN using pseudo-labels in weakly-supervised segmentation tasks~\cite{bansal2017pixelnet,khoreva2017simple,li2018weakly,ahn2019weakly,ahn2018learning}; we took this approach for our tracking task.
We train a state-of-the-art cell tracking network called MPM-Net~\cite{hayashida2019mpm} with the pseudo-labels.
The MPM-Net estimates Motion and Position Map (MPM) that simultaneously represents the cell positions and their association between frames from inputting the images at successive frames.
The advantage of this method is that it can extract the features of the spatio-temporal context about nearby cells from the inputted entire images for association and detection.
We generate the pseudo-training data for MPM using obtained cell position and association in Sec. \ref{sec:BFProp}.

\begin{figure}[t]
    \centering
    \begin{tabular}{c}
      \begin{minipage}{0.48\textwidth}
      \centering
        \includegraphics[width=0.95\linewidth]{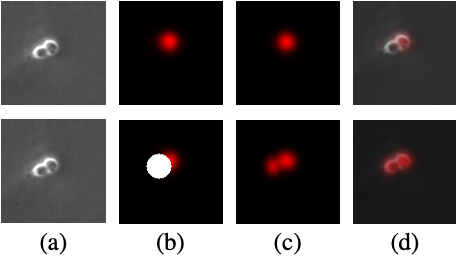}
        \vspace{-2mm}
        \caption{Example result using masked loss. (a) phase-contrast image, (b) generated pseudo-training data, (c) output of trained network, and (d) overlapping images. The top row is an example not using the masked loss. The bottom row is an example using the masked loss. The white region in (b) indicates the region that ignores the loss.
        }
        \label{fig:mask loss}
      \end{minipage}%
      
      \begin{minipage}{0.03\textwidth}
      \hspace{0.5mm}
      \end{minipage}%
      
      \begin{minipage}{0.48\textwidth}
      \centering
        \includegraphics[width=1\linewidth]{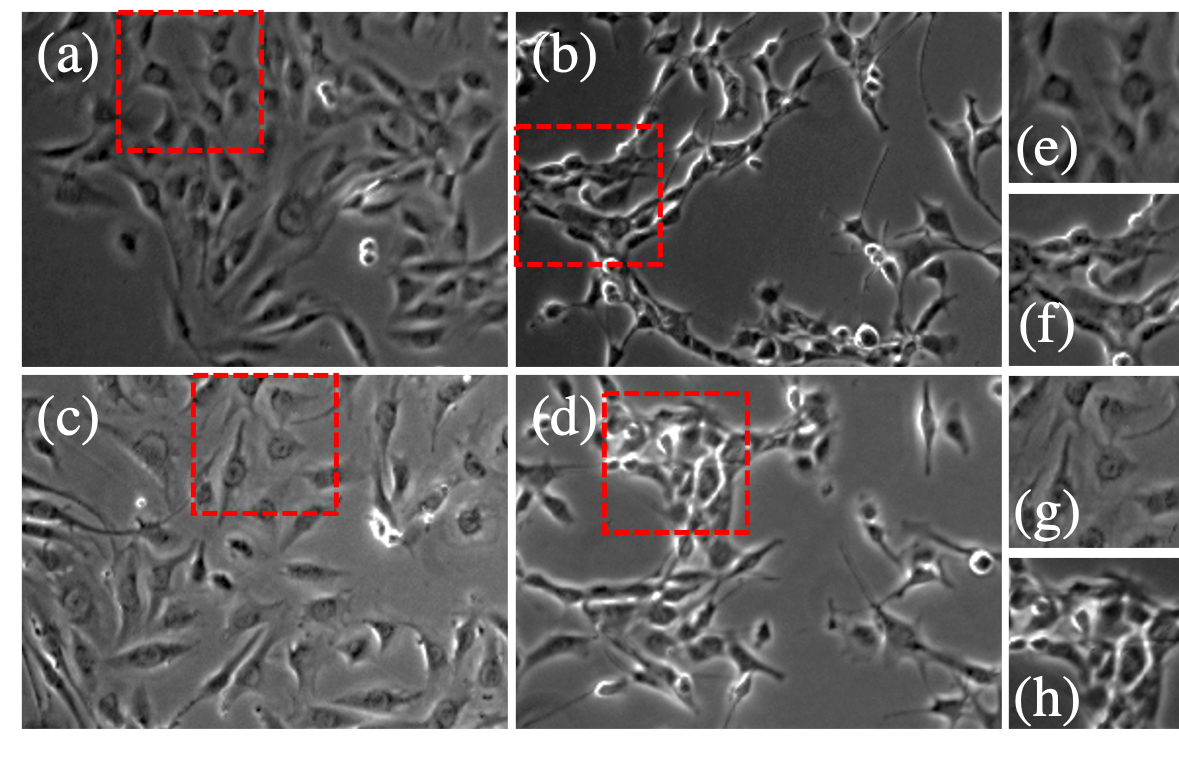}
        \vspace{-7mm}
        \caption{Example images on four calture condtions. (a) Control, (b) FGF2, (c) BMP2, (d) FGF2+BMP2, (e)-(h) are enlarged images of the red box in (a)-(d). The appearance of the cells depend on the culture conditions. Under the FGF2 and FGF2+BMP2 conditions, the cell often shrink and overlap.}
        \label{fig:dataset}
      \end{minipage}
    \end{tabular}
    \vspace{-5mm}
\end{figure}

In the previous step, we obtained the high confidence pseudo-labels and a set of unassociated regions $\mathbf{\Gamma}$.
The top row in Fig.~\ref{fig:mask loss}(b) shows an example of pseudo-training data directly generated using only the high confidence labels. 
In this example, a cell divides two cells. However, the mother cell was associated with one of the child cells and the other was not (false negative) due to one-by-one matching. If we train the network using such noisy labels, this non-associated cell region affects the learning. Indeed, the non-associated region was not detected due to over-fitting.
Fig.~\ref{fig:mask loss}(c) shows the output of the trained network that only detects one cell by over-fitting. 
To avoid this problem, we train the network with the masked loss function that ignores the loss from the false-negative regions where the cell did not correspond to any detection responses due to its low confidence.
The masked loss is formulated as:
\begin{eqnarray}
    Loss_{mask} = 
    \begin{cases}
        0 &\mbox{if} ~\bm{p} \in \mathbf{\Gamma}, \\
        Loss_{ori} &\mbox{otherwise},
    \end{cases}
\end{eqnarray}
where $\mathbf{\Gamma}$ is the set of the ignoring regions that contain the unassociated cell regions, $\bm{p}$ is a pixel in $\mathbf{\Gamma}$, $Loss_{ori}$ is the original loss for MPM-Net~\cite{hayashida2019mpm}.
As shown in the white circle in bottom row of Fig.~\ref{fig:mask loss}(b), the false-negative region is not calculated in the masked loss. This effectively avoid over-fitting and correctly estimate the cell position likelihood map as shown in the bottom row of Fig.~\ref{fig:mask loss}(c).

\vspace{-2mm}
\section{Experiment}
\vspace{-3mm}
\subsection{Data set and experimental setup}
\vspace{-3mm}
We evaluated our method on an open data set~\cite{elmer} that contains time-lapse sequences captured phase-contrast microscopy\footnote{The data~\cite{elmer} is more challenging as a tracking task compared with ISBI Cell Tracking Challenging~\cite{ulman2017objective,mavska2014benchmark} that more focused on segmentation task, since the cells often partially overlapped and the boundary of cells is ambiguous.}.
In the data set, the mybolast cells were cultured under four growth factor conditions: (a) Control, (b) FGF2, (c) BMP2, and (d) FGF2+BMP2 (Fig.~\ref{fig:dataset}). Each sequence consists of 780 frames, with a 5 minute interval between consecutive frames. The resolution of each image is 1392$\times$ 1040 pixels.
There are four sequences for each condition and the total number of sequences is 16.
The rough cell centroid positions are annotated with the cell ID.
In one of the BMP2 sequence, all cells are annotated.
For the other sequences, three cells were randomly selected at the beginning of the sequence and then their descendants were annotated.
The total number of annotated cells in the 16 sequences is 135859.
We used one of the BMP2 sequence as the training data for co-detection CNN and the other sequences were used as the test data. 
In the training process, we only used the cell position coordinates as weak labels.
The task was challenging because the training was only weakly-supervised and the appearances of the cells in the test data differed from those in the training data (see Fig.~\ref{fig:dataset}).
To train co-detection CNN and MPM, we used Adam~\cite{kingma2014adam} optimizer with learning rate $10^{-3}$.
We set the threshold $th$ in Eq.~\ref{eq:forwardmask} to 0.01, the low confidence association threshold $th_{conf}$ to 0.5, and $r=18$ in all experiments; these parameters were decided using validation data and were not sensitive. 

\vspace{-3mm}
\subsection{Performance of cell tracking on open data set}
We compared our method with five other methods by using the open data set~\cite{elmer}. 
Since our method only requires weak-supervision, we selected three methods that do not use association information;
1) asymmetric graphcut (A-Graph)~\cite{BenschR2015} that segments cell regions using asymmetric graph-cut: it was trained with the small amount of additional ground-truth for segmentation;
2) Fogbank~\cite{ChalfounJ2016} that segments cell regions using image processing: the hyper-parameters were tuned using the validation data; 3) global data association (GDA)~\cite{bise2011reliable} that segments cell regions by physical-model-based method~\cite{yin2012understanding} and then performs spatial-temporal global data association: the hyper-parameters were tuned using validation data.
In addition, in order to show that the performance of our method is comparable with the SOTA  (state-of-the-art), we evaluated two supervised tracking methods that require the ground-truth of the cell position and association; 4) cell motion field (CMF)~\cite{hayashida2019cell} that estimates the cell motion and position separately; 5) motion and position map (MPM)~\cite{hayashida2019mpm} that estimates the motion and position map, which achieved the SOTA performance.
In addition, to confirm the effectiveness of the masked loss, we also compared with our method without the masked loss (Ours w/o ml).

We used the association accuracy and target effectiveness as following the paper that proposed the MPM~\cite{hayashida2019mpm}. 
Each target was first assigned to a track (estimation) for each frame.
The association accuracy indicates the number of true positive associations divided by the number of true positive associations in the ground-truth. If cell A switches into B, and B into A, we count two false-positive (A$\rightarrow$B, B$\rightarrow$A) and two false-negatives (no A$\rightarrow$A, B$\rightarrow$B).
The target effectiveness was computed as the number of the assigned track observations over the total number of frames of the target after assigning each target to a track that contains the most observations from that ground-truth.
It indicates how many frames of targets are followed by
computer-generated tracks. This metric is a stricter than the association accuracy.
If a switching error occurs in the middle of the trajectory, the target effectiveness is 0.5.

\begin{figure}[t]
    \centering
    \begin{tabular}{ccc}
      \begin{minipage}{0.48\textwidth}
      \centering
        \includegraphics[width=\linewidth]{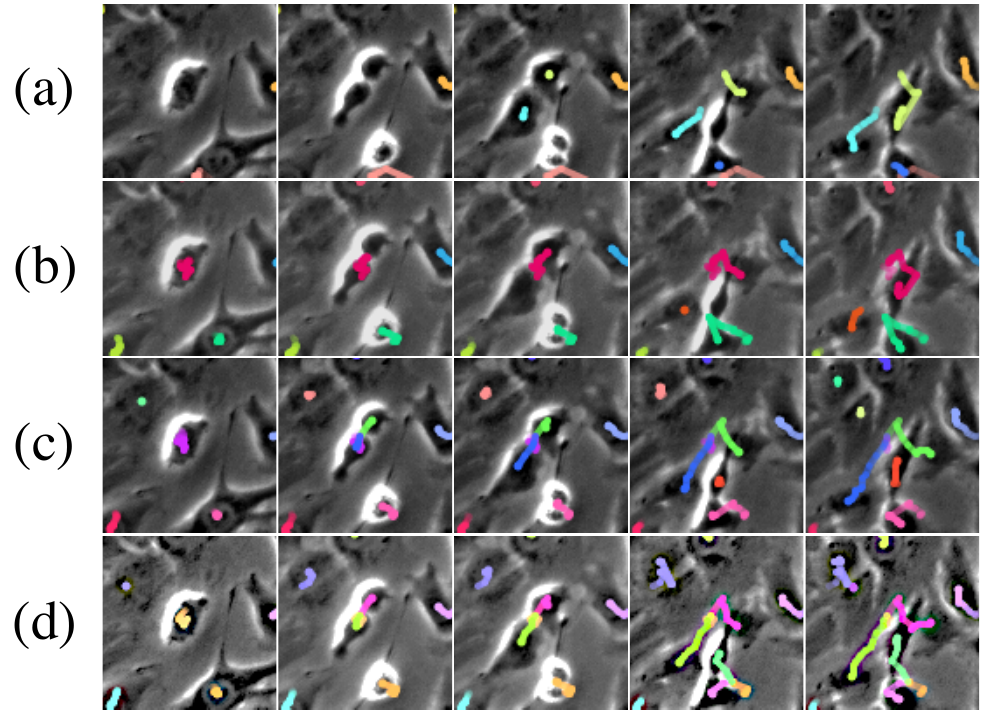}
        \vspace{-7mm}
        \caption{Examples of tracking results of (a) GDA, (b) CMF, (c) MPM, and (d) ours. The horizontal axis indicates the time.}
        \label{fig:compare with others}
      \end{minipage}%
      
      \begin{minipage}{0.03\textwidth}
      \hspace{0.5mm}
      \end{minipage}%
      
      \begin{minipage}{0.48\textwidth}
      \centering
        \includegraphics[width=\linewidth]{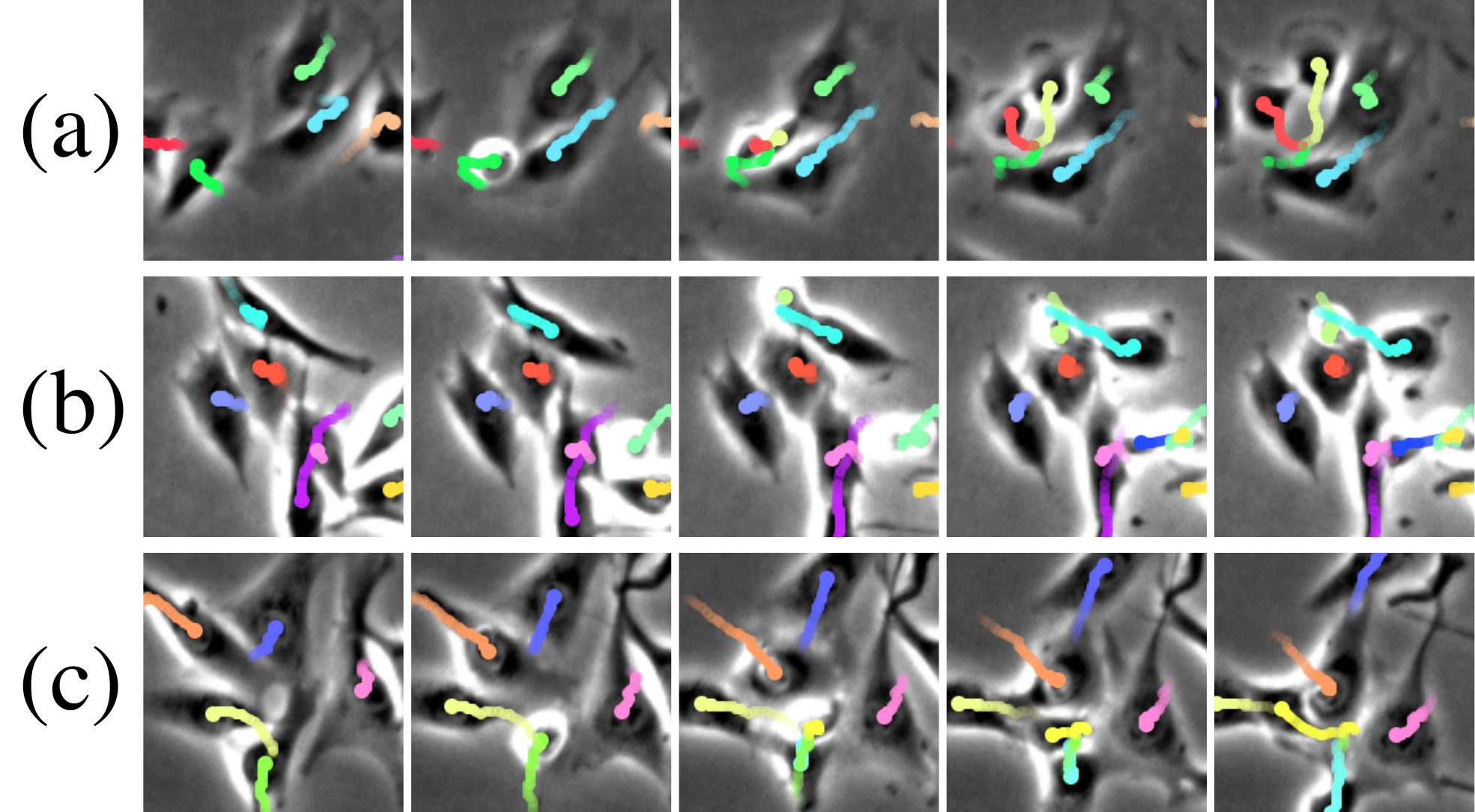}
        \vspace{-5mm}
        \caption{Examples of tracking results under each conditions: (a) Control, (b)FGF2, and (c) FGF2+BMP2. }
        \label{fig:various condtions}
      \end{minipage}
    \end{tabular}
\end{figure}

Table~\ref{tab:evaluation} shows the results of the performance comparison.
Our method outperformed the other weakly or unsupervised methods (A-Graph~\cite{BenschR2015}, Fogbank~\cite{ChalfounJ2016}, GDA~\cite{bise2011reliable}) and achieved comparable results with state-of-the-art supervised methods (MPM~\cite{hayashida2019mpm}).
Even though our method used only weak-supervision, it outperformed that of the supervised MPM in FGF2. We consider that the MPM may be over-fitted to the condition of the training data (BMP2), and thus its performance may decrease since the cell appearance in FGF2 is different from that in BMP2 as shown in Fig.~\ref{fig:dataset}. In addition, the results show that the masked loss slightly improved the performance compared with 'Ours w/o ml'.
Fig.~\ref{fig:compare with others} (c) shows examples of tracking results under BMP2\footnote{Since the tracking results of A-Graph and Fogbank were very worse, we omitted their results on these figures due to the page limitation.}. Although GDA did not detect the brighter cell and CMF did not identify the newly born cells after cell mitosis, our method successfully tracked almost all the cells as the same with MPM.

\begin{table}[t]
\centering
\caption{Tracking performance in terms of association accuracy (AA) and target effectiveness (TE) using open data set~\cite{elmer}. Su. indicates the condition of the training data: weak-supervision (W), un-supervision (U), and fully-supervision (F). The best and second best are denoted by boldface and the best one is underlined. '*' indicates the culture condition is the same as in the training data; no asterisk means the culture conditions were different.
}
\vspace{-2mm}
\label{tab:evaluation}
\small
\begin{tabular}{c|c|cc|cc|cc|cc|cc}
\hline \multirow{2}{*}{Method} &\multirow{2}{*}{Su.} & \multicolumn{2}{c}{*BMP2} & \multicolumn{2}{c}{FGF2} & \multicolumn{2}{c}{Control} & \multicolumn{2}{c}{\begin{tabular}[c]{@{}c@{}}FGF2+\\ BMP2\end{tabular}} & \multicolumn{2}{c}{Ave.} \\ 
                                                & & AA & TE & AA & TE & AA & TE & AA & TE & AA & TE \\ \hline
            \itshape A-Graph~\cite{BenschR2015} & W & 0.801& 0.621& 0.604& 0.543& 0.499& 0.448 & 0.689& 0.465& 0.648&0.519\\
            \itshape Fogbank~\cite{ChalfounJ2016}& U & 0.769& 0.691& 0.762& 0.683& 0.650& 0.604&  0.833& 0.587& 0.753& 0.641\\
            \itshape GDA~\cite{bise2011reliable} & U & 0.855& 0.788& 0.826& 0.733& 0.775& 0.710&  0.942& 0.633& 0.843& 0.771\\
            \itshape Ours w/o ml & W & 0.979& 0.960 & 0.950& 0.861& 0.917& 0.786&  0.972& 0.880& 0.954& 0.873\\ 
            \itshape Ours & W & \textbf{0.982}& \textbf{\underline{0.970}} & \textbf{\underline{0.955}}& \textbf{\underline{0.869}}& \textbf{0.926}& \textbf{0.806}& \textbf{0.976}& \textbf{\underline{0.911}}& \textbf{0.960}& \textbf{\underline{0.881}}\\ \hline
            \itshape CMF~\cite{hayashida2019cell} & F & 0.958& 0.939 & 0.866& 0.756& 0.884& 0.761& 0.941&0.841& 0.912& 0.822\\
            \itshape MPM~\cite{hayashida2019mpm} & F  & \textbf{\underline{0.991}}& \textbf{0.958}& \textbf{0.947}& \textbf{0.803}& \textbf{\underline{0.952}}& \textbf{\underline{0.829}}& \textbf{\underline{0.987}}& \textbf{\underline{0.911}}& \textbf{\underline{0.969}}& \textbf{0.875}\\ \hline
\end{tabular}
\vspace{-3mm}
\end{table}

\subsection{Ablation study}
Next, we performed an ablation study to evaluate the performance of the co-detection CNN, backward-and-forward propagation, and re-training individually. 
In this ablation study, in order to confirm the robustness in various conditions, we additionally added annotations for other three conditions (Control, FGF2, FGF2+BMP2) since only some of the cells were annotated under these three conditions in the original data. Then, we evaluated each step of our method under these three conditions.

\begin{figure}[t]
  \centering
  \begin{tabular}{ccc}
  \begin{minipage}{0.3\hsize}
    \centering
    \includegraphics[width=\linewidth]{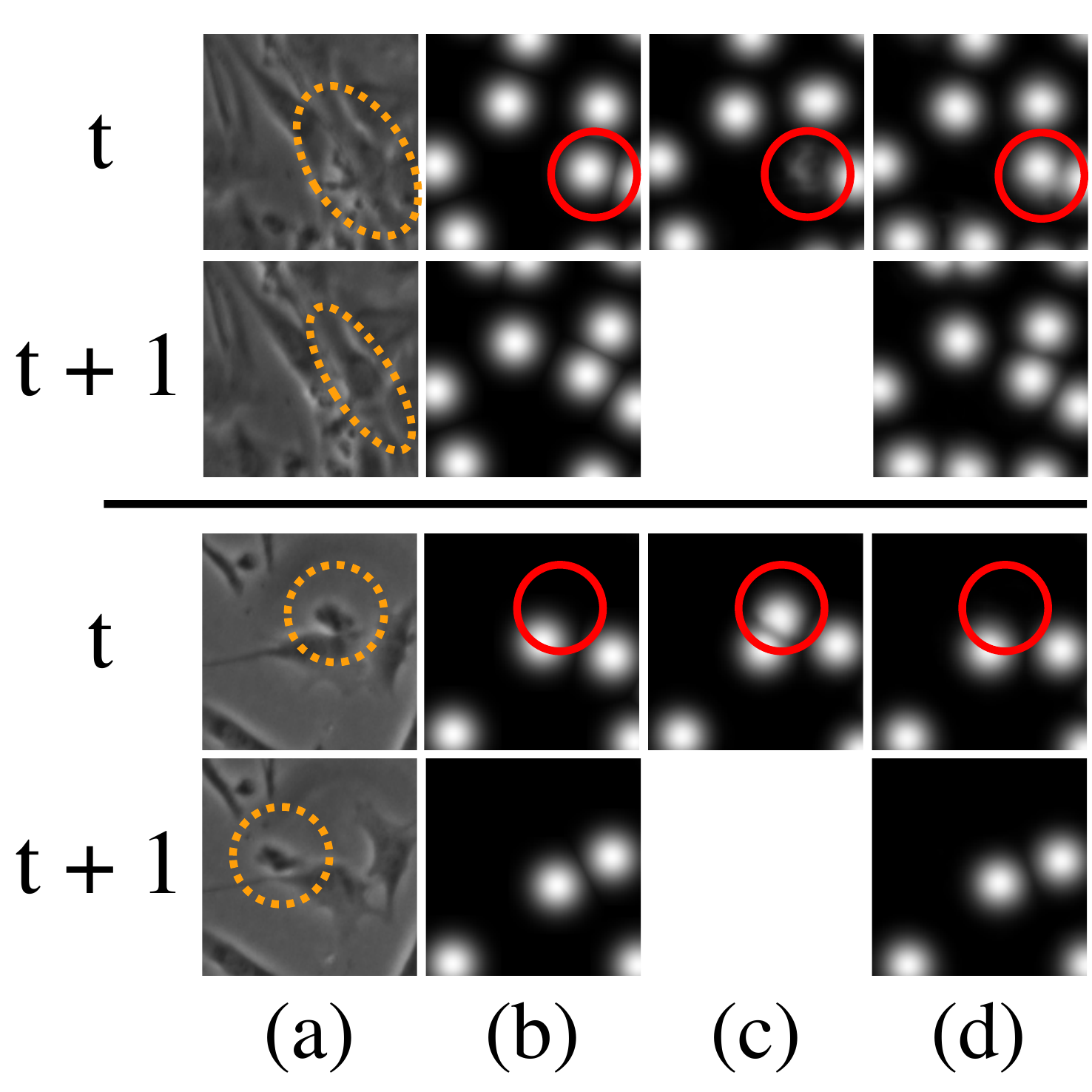}
    \caption{Examples results of tracking. (a) phase-contrast, (b) ground-truth, (c) Nishimura~\cite{nishimura2019weakly}, and (d) co-detection CNN.}
    \label{fig:example_detection}
  \end{minipage}%
  
  \begin{minipage}{0.02\hsize}
    \hspace{1mm}
  \end{minipage}%
  
  \begin{minipage}{0.3\hsize}
    \centering
    \includegraphics[width=\linewidth]{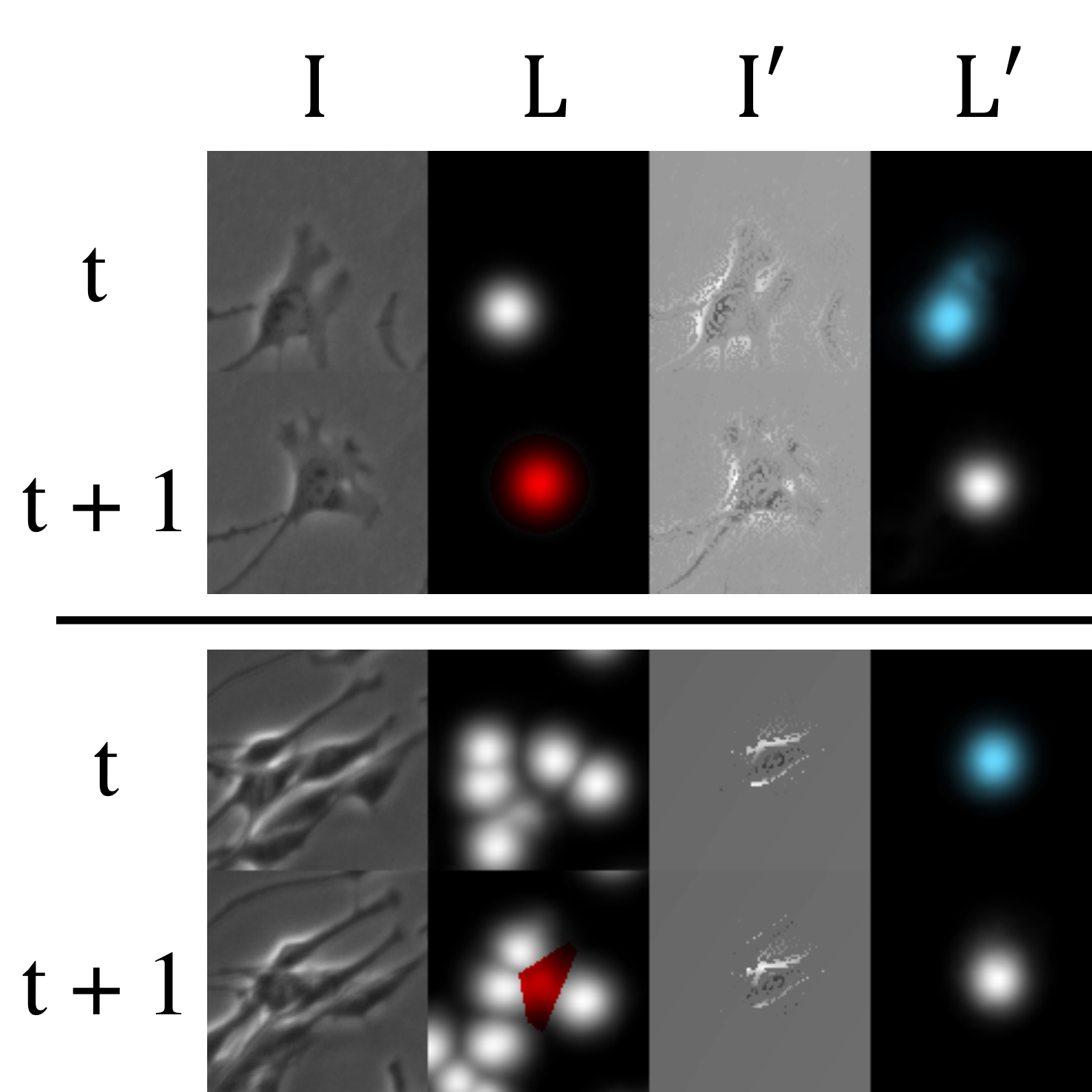}
    \caption{Example results of BF-prop in (a) simple case and (b) complex case. The red regions are the cell region of interest, and the blue regions are the estimated corresponding cell.}
    \label{fig:Backwardandforward}
  \end{minipage}
  
  \begin{minipage}{0.02\hsize}
    \hspace{1mm}
  \end{minipage}%
  
  \begin{minipage}{0.35\textwidth}
      \centering
        \includegraphics[width=0.9\linewidth]{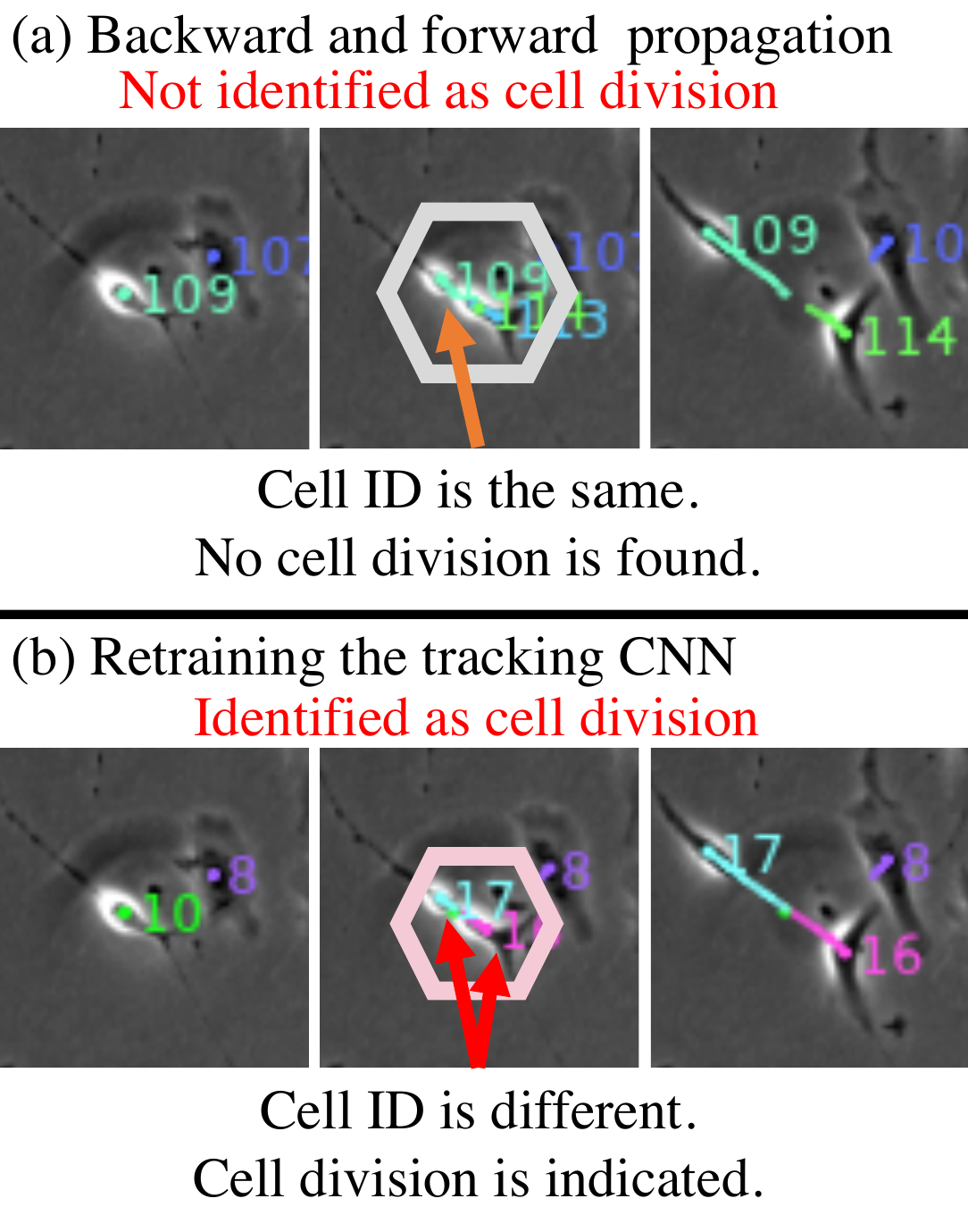}
        \vspace{-5mm}
        \caption{Example results from (a) BF-prop, (b) retrained MPM-net.  
        }
        \label{fig:Bfwithtrain}
      \end{minipage}
  \end{tabular}
  \vspace{-1mm}
\end{figure}

\noindent{\bf Co-detection CNN:}
We first evaluated our co-detection CNN against the method proposed by Nishimura {\it et al.}~\cite{nishimura2019weakly} that estimates the cell position likelihood map of a single image. In addition, in order to demonstrate the robustness for cell migration speed since the speed is depending on the cell types and time-interval, we also evaluated two intervals (5 and 25 minutes), in which the speed in 25 min is much faster than that in 5 min.
We used F1-score as the detection performance metric.
\begin{table}[t]
    \centering
    \caption{Detection performance.}
    \begin{tabular}{c|cccccc}
    \hline Method & \begin{tabular}[c]{@{}c@{}}*BMP2\\ sparse\end{tabular} & \begin{tabular}[c]{@{}c@{}}*BMP2\\ medium\end{tabular} & \begin{tabular}[c]{@{}c@{}}*BMP2\\ dense\end{tabular} & Control & FGF2  & \begin{tabular}[c]{@{}c@{}}FGF2\\+BMP2\end{tabular}\\ \hline 
    \itshape Nishimura~\cite{nishimura2019weakly} & 0.998  & 0.978  & 0.977 & 0.922 & 0.924 & \textbf{0.962}     \\
    \itshape Ours (5 min. int.)       & \textbf{0.999}  & 0.983  & \textbf{0.980} & 0.923 & \textbf{0.928} & 0.945     \\
   \itshape Ours (25 min. int.)      & 0.998  & \textbf{0.984}  & 0.978 & \textbf{0.926} & 0.911 & 0.946    \\ \hline
    \end{tabular}
    \label{tab:co-detection_result}
\end{table}
Table~\ref{tab:co-detection_result} shows the results.
Our co-detection CNN performed almost as well as the state-of-the-art method under all conditions (BMP2-sparse, BMP2-medium, BMP2-dense, Control, FGF2, FGF2+FGF2). In addition, the results show that our method was robust to the different cell migration speeds, since the performances of both interval conditions are almost the same.
Fig.~\ref{fig:example_detection} shows examples in which co-detection CNN improved the detection results. In the upper case, the cell shape is ambiguous at $t$ but it is more clear at $t+1$, co-detection CNN uses these two image and it may facilitate to detect the cell. In the bottom case, a tips of cell (noise) appears at both frames. Since the noise traveled a large distance, co-detection CNN may reduce over-detections.

\begin{table}[t]
    \centering
    \caption{Association performance of backward-and-forward propagation.}
    \vspace{-3mm}
    \begin{tabular}{c|c|cccccc}
    \hline 
    Interval & Metrics & \begin{tabular}[c]{@{}c@{}}*BMP2\\ sparse\end{tabular} & \begin{tabular}[c]{@{}c@{}}*BMP2\\ medium\end{tabular} & \begin{tabular}[c]{@{}c@{}}*BMP2\\ dense\end{tabular} & Control & FGF2  & \begin{tabular}[c]{@{}c@{}}FGF2 \\+BMP2\end{tabular}\\ \hline 
    \multirow{3}{*}{5 min.} & Precision  &
    0.999 &0.989&  0.992& 0.971& 0.966&  0.964    \\
    & Recall    & 0.997&  0.976&  0.957& 0.849& 0.844&  0.900     \\
    & F1-score      & 0.998&  0.982&  0.974& 0.906& 0.901&  0.931     \\ \hline
    \multirow{3}{*}{25 min.} & Precision     & 0.998&  0.982&  0.974& 0.906& 0.901&  0.931    \\
    & Recall  & 0.961& 0.975& 0.960& 0.849& 0.753& 0.902    \\
    & F1-score   & 0.979& 0.981& 0.976& 0.904& 0.830& 0.930    \\ \hline
    \end{tabular}
    \label{tab:BFprop}
    \vspace{-3mm}
\end{table}

\noindent{\bf Backward-and-forward propagation:}
Next, we evaluated the association performance of Backward-and-Forward propagation (BF-prop).
We used precision, recall, F1-score of association accuracy as the performance metrics. 
Fig.~\ref{fig:Backwardandforward} shows examples. $L$ indicates the output of co-detection CNN given two input images I at $t$ and $t+1$. I' indicates the masked image generated using the relevance map produced by backward propagation. $L'$ indicates the estimated likelihood map by forward propagation by inputting $I'$. In both cases, the backward propagation could obtain the target cell regions and forward propagation successfully estimated the detection map of the corresponding cell.
Under all conditions, backward-and-forward propagation performed association accurately (over 90\% in the terms of F1-score as shown in Table~\ref{tab:BFprop}).
As discussed in Sec. \ref{sec:BFProp}, precision is more important than recall when using the pseudo-labels. BF-prop achieved higher precision than recall on all data-sets. In addition, we conducted evaluations using the different intervals (5 and 25 min.). The results for 5 min. were slightly better than those in 25 min, but not significantly.

\begin{table}[t]
\centering
\caption{Comparison of backward-and-forward propagation (BF) with MPM trained by backward-and-forward propagation (T). AA: association accuracy, TE: target effectiveness.}
\label{tab:retrain result}
\footnotesize
\begin{tabular}{c|cc|cc|cc|cc|cc|cc}
    \hline 
    \multirow{3}{*}{Met.} & \multicolumn{2}{c}{*BMP2}  & \multicolumn{2}{c}{*BMP2}  & \multicolumn{2}{c}{*BMP2}  & \multicolumn{2}{c}{\multirow{2}{*}{Control}} & \multicolumn{2}{c}{\multirow{2}{*}{FGF2}}  & \multicolumn{2}{c}{FGF2}\\
    &  \multicolumn{2}{c}{sparse} & \multicolumn{2}{c}{medium} & \multicolumn{2}{c}{dense} &  &  & & &\multicolumn{2}{c}{+BMP2} \\ 
    & AA & TE & AA & TE & AA & TE & AA & TE & AA & TE & AA & TE \\ \hline
    \itshape BF & 0.983& 0.968	& 0.973&	0.962 & 0.840	& 0.914 &	0.826&	0.765&	\textbf{0.794} & \textbf{0.672} &	0.955 &	0.945  \\ 
    \itshape T &  \textbf{0.993} &\textbf{0.976}&	\textbf{0.980} & \textbf{0.974}& \textbf{0.970} & \textbf{0.969}&	\textbf{0.858} & \textbf{0.800} & 0.773 & 0.640 & \textbf{0.982} & \textbf{0.970} \\ \hline
\end{tabular}
\vspace{-2mm}
\end{table}

\noindent{\bf Training tracking CNN:}
To show the effectiveness of retraining the tracking CNN, we compared the trained CNN with the results of BF-prop in terms of the same tracking metrics (AA and TE).
As shown in Table~\ref{tab:retrain result}, the retraining improved the performance under almost conditions except FGF2.
The important thing is that although the BF-prop only tracks one of the cell when a cell divided two cells, the retraining could identify a cell division since the masked loss helped to detect the another cell of the divided two cells. Fig.~\ref{fig:Bfwithtrain} shows the example of the cell division case.
The two new cells were successfully identified by MPM-Net and new IDs were assigned to them. In contrast, BF-prop tracked only one of them continuously and did not identify cell division.

\begin{figure}[t]
    \centering
    \begin{tabular}{c}
      \begin{minipage}{0.65\hsize}
        \centering
        \includegraphics[width=\linewidth]{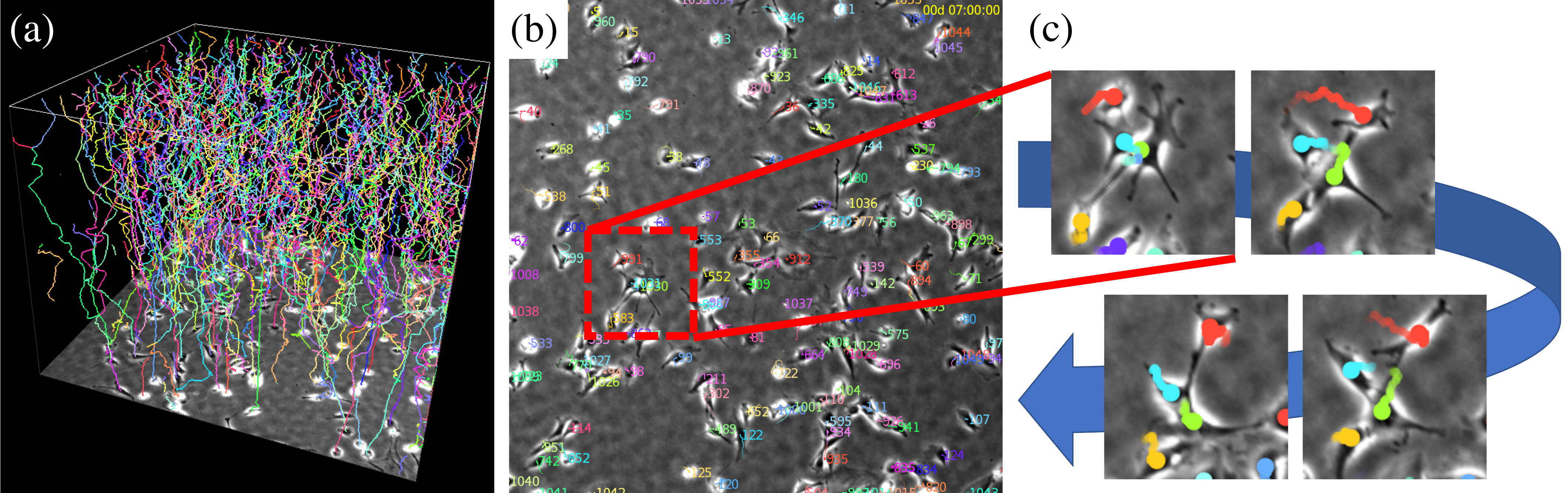}
        \caption{Examples of tracking result on real data. (a) 3D view of estimated cell trajectories. The z-axis is the time, and each color indicates the trajectory of a cell. (a) Entire image. (c) Sequence of enlarged images of image sequence at the red box in (b).
        }
        \label{fig:elmer}
      \end{minipage}
      
      \begin{minipage}{0.03\hsize}
      \hspace{0.5mm}
      \end{minipage}
      
      \begin{minipage}{0.29\hsize}
        \centering 
        \tblcaption{Quantitative evaluation. AA: association accuracy, TE: target effectiveness}
        \begin{tabular}{ccc}
        \hline Method   &  AA   & TE   \\ \hline
        \itshape A-Graph~\cite{BenschR2015}  & 0.216 & 0.169 \\
        \itshape Fogbank~\cite{ChalfounJ2016} & 0.695 & 0.321  \\
        \itshape GDA~\cite{bise2011reliable} & 0.773 & 0.527  \\ 
        \itshape Ours & \textbf{0.857} & \textbf{0.804}   \\ \hline
        \end{tabular}\label{tab:elmer eval}
      \end{minipage}
    \end{tabular}
    \vspace{-3mm}
\end{figure}

\subsection{Cell tracking without any human annotation}
In this section, we consider a more realistic scenario when pairs of phase-contrast and fluorescent microscopy images for training and the test image sequence that was captured by only phase-contrast microscopy were provided by biologists without any human annotation.
In order to confirm that our method can perform such a realistic scenario, we also prepared a data-set with this problem setup.
In this experiment, 86 pairs of phase-contrast and fluorescent images were given as the training data, and the 95 images was given as the test data, in which the cell appearance is different from the open data set we used in the previous section.

In this setting, our method could perform tracking in four steps.
(1) We trained the detection CNN with phase-contrast images using the ground-truth of detection automatically generated from the given fluorescent image(the procedure was the same as that of Nishimura~\cite{nishimura2019weakly}.). Then, we generated co-detection pseudo labels.
(2) We trained co-detection CNN with the generated detection pseudo labels, and
(3) generated the pseudo-labels for association by BF-prop.
(4) We trained the MPM-net using the pseudo-labels and applied it to the test data.
In the evaluation, we also compared our method with un-supervised and weakly-supervised tracking methods; A-Graph~\cite{BenschR2015}, Fogbank~\cite{ChalfounJ2016}, GDA~\cite{bise2011reliable}. Here, we could not compare with the supervised method on this scenario since there was no supervised training data.

Table~\ref{tab:elmer eval} shows the tracking results in terms of association accuracy (AA) and target effectiveness (TE). Our method outperformed the other methods on both metrics.
Our method achieved an 8\% improvement in association accuracy and 28\% improvement in target effectiveness compared with the second best.
As shown in Fig.~\ref{fig:elmer}, our method can track many cells without supervised annotation.
These results show that our method can effectively use weak labels and obtain good tracking results.

\section{Conclusion}
We proposed a weakly-supervised tracking method that can track multiple cells by using only training data for detection. 
The method first trains co-detection CNN that detects cells in successive frames by using weak supervision. Then, the method obtains association from co-detection CNN by using our novel backward-and-forward propagation method on the basis of the key assumption that co-detection CNN implicitly learns the association. The association 
is used as pseudo-labels for a state-of-the-art tracking network (MPM-net). 
Our method outperformed the compared methods and achieved comparable results to those of supervised state-of-the-art methods.
In addition, we demonstrated the effectiveness of our method in a realistic scenario in which the tracking network was trained without any human annotations.

\vspace{4mm}
\noindent
\textbf{Acknowledgement:}
This work was supported by JSPS KAKENHI Grant Number 20H04211.

\bibliographystyle{splncs04}
\bibliography{egbib}
\end{document}